\documentclass[runningheads]{llncs_short_figure_margin}
\usepackage{graphicx}

\usepackage{tikz}
\usepackage{comment}
\usepackage{amsmath,amssymb} 
\usepackage{color}
\usepackage{wrapfig,lipsum,booktabs}
\usepackage{hyperref}       
\usepackage{url}            
\usepackage{booktabs}       
\usepackage{amsfonts}       
\usepackage{nicefrac}       
\usepackage{microtype}      
\usepackage{xcolor}         
\usepackage[numbers,comma,sort]{natbib}
\usepackage{multirow}
\usepackage{gensymb}
\usepackage{subfiles}
\usepackage{xr}
\usepackage{wrapfig}
\usepackage{footmisc}
\usepackage{float}

\usepackage[numbers,comma,sort]{natbib}
\usepackage{hyperref}
\usepackage{booktabs}
\usepackage{multirow}
\usepackage{enumitem}
\usepackage{comment}
\usepackage{caption}
\usepackage{tabularx}
\usepackage{amsmath}
\usepackage{amssymb}

\hypersetup{
    colorlinks=true,
    linkcolor=blue,
    filecolor=magenta,      
    urlcolor=cyan,
    pdfpagemode=FullScreen,
}

\let\oldmaketitle\maketitle
\renewcommand{\maketitle}{\oldmaketitle\setcounter{footnote}{0}}

\newcommand{\ourmethod}{MVDiffusion++}

\newcommand\mypara[1]{\vspace{0.2cm}\noindent\textbf{#1.}}
\newcommand\myparapara[1]{\vspace{0.2cm}\noindent\textbf{#1}}

\newcommand{\yasu}[1]{{\color{cyan}{\bf [Yasu: #1]}}}

\newcolumntype{C}{>{\raggedright\arraybackslash}X} 

\newcommand{\expec}{\mathbb{E}}

\newcommand{\hpixel}{H}
\newcommand{\wpixel}{W}
\newcommand{\cpixel}{3}

\newcommand{\hlatent}{h}
\newcommand{\wlatent}{w}
\newcommand{\clatent}{c}

\newcommand{\lsimpleldm}{L_{LDM}}

\newcommand{\lsimplemv}{L_{MVLDM}}

\newcommand{\model}{\epsilon_\theta}
\newcommand{\conditioner}{\lambda_\theta}

\newcommand{\encoder}{\mathcal{E}}
\newcommand{\decoder}{\mathcal{D}}

\newcommand{\bx}{\mathbf{x}}
\newcommand{\by}{\mathbf{y}}

\newcommand{\bepsilon}{{\boldsymbol{\epsilon}}}
\newcommand{\bv}{{\boldsymbol{v}}}

\newcommand{\img}{\bx}
\newcommand{\imgrec}{\tilde{\img}}

\newcommand{\latent}{Z}

\newcommand{\cond}{\by}

\newcommand\orange[1]{{\color{orange}#1}}

\usepackage[accsupp]{axessibility}  

\usepackage[width=122mm,left=12mm,paperwidth=146mm,height=193mm,top=12mm,paperheight=217mm]{geometry}

\newcommand{\dilin}[1]{\textcolor{blue}{\small dilin: #1}}

\begin{document}
\pagestyle{headings}
\mainmatter
\def\ECCVSubNumber{100}  

\title{MVDiffusion++: A Dense High-resolution Multi-view Diffusion Model for Single or Sparse-view 3D Object Reconstruction}

\titlerunning{\ourmethod}
%
\author{Shitao Tang\inst{1*} \and
Jiacheng Chen\inst{1*} \and
Dilin Wang\inst{2*} \and Chengzhou Tang\inst{2} \and \\ Fuyang Zhang\inst{1} \and Yuchen Fan\inst{2} \and Vikas Chandra\inst{2}  \and \\ Yasutaka Furukawa\inst{1\dag} \and Rakesh Ranjan\inst{2\dag}}

\authorrunning{Tang et al.}
%
\institute{Simon Fraser University \\ \email{\{shitaot, jca348, fuyangz, furukawa\}@sfu.ca} \and Meta Reality Labs \\ \email{\{wdilin, chengzhout, ycfan, vchandra, rakeshr\}@meta.com}}

\maketitle
\let\thefootnote\relax\footnote{*Equal contribution. \dag Joint last author. 
Work done during an internship with Meta. 
}

\begin{abstract}
This paper presents a neural architecture {\ourmethod}~for 
3D object reconstruction that synthesizes dense and high-resolution views of an object given one or a few images without camera poses.
\ourmethod~achieves superior flexibility and scalability with two surprisingly simple ideas: 1) A ``pose-free architecture'' where standard self-attention among 2D latent features learns 3D consistency across an arbitrary number of conditional and generation views without explicitly using camera pose information; and 2) A ``view dropout strategy'' that discards a substantial number of output views during training, 
which reduces the training-time memory footprint and enables dense and high-resolution view synthesis at test time.
We use the Objaverse for training and the Google Scanned Objects for evaluation with standard novel view synthesis and 3D reconstruction metrics, where \ourmethod~significantly outperforms the current state of the arts. We also demonstrate a text-to-3D application example by combining {\ourmethod} with a text-to-image generative model.
The project page is at \url{https://mvdiffusion-plusplus.github.io}.

\end{abstract}

\section{Introduction}

\begin{figure}[t]
\centering
\includegraphics[width=\linewidth]{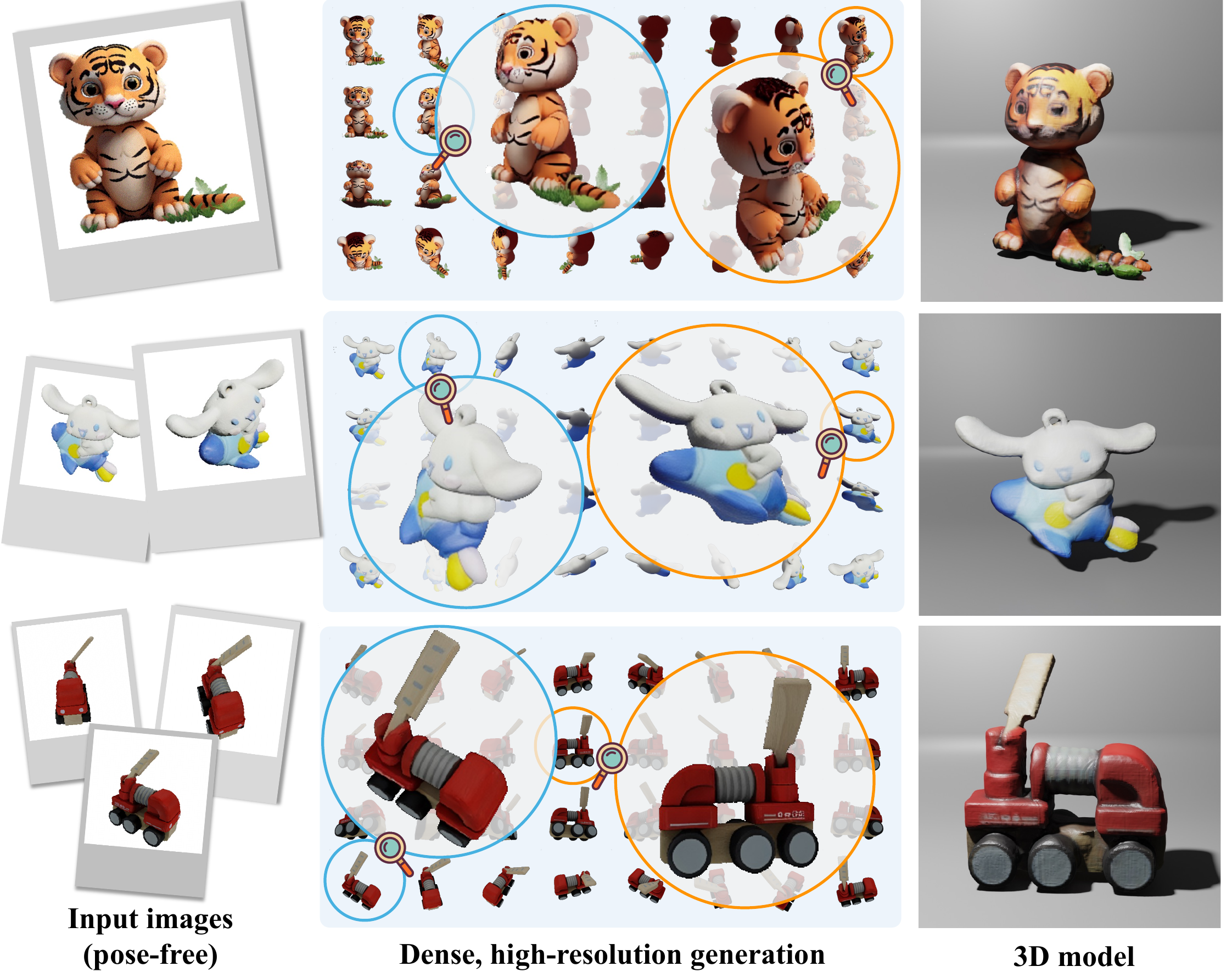}
\caption{\ourmethod~generates dense(32) and high-resolution(512$\times$512) images of an object from a single or multiple unposed images. 
The input images of the three examples are from 
a latent diffusion model, OmniObject3D\cite{wu2023omniobject3d}, and Google Scanned Objects\cite{downs2022gso}, respectively. 
}
\label{fig:teaser}
\end{figure}



Human vision demonstrates remarkable flexibility. Look at the images of objects at the left in \autoref{fig:teaser}. 
While unable to create millimeter-accurate 3D models, our visual system can combine information from a few images to form a coherent 3D representation in our minds, including intricate facial features of a tiger or the arrangement of blocks forming a toy train, even parts that are fully obscured.

3D reconstruction technology~\cite{stereopsis2010accurate, yao2018mvsnet, agarwal2011building, furukawa2010towards} has evolved over the last fifteen years in a fundamentally different way. Unlike the human ability to infer 3D shapes from a few images, the technology takes hundreds of images of an object, estimates their precise camera parameters, and reconstructs high-fidelity 3D geometry at a sub-millimeter accuracy. 

This paper explores a new paradigm of 3D reconstruction that combines the high-fidelity of computational methods and the flexibility of human visual systems. Our inspiration comes from exciting recent developments in multi-view image generative models~\cite{tang2023mvdiffusion,shi2023mvdream,liu2023syncdreamer,long2023wonder3d,shi2023zero123++,liu2023one2345++, wang2023imagedream}. MVDiffusion~\cite{tang2023mvdiffusion} is an early attempt to extend pre-trained image diffusion models~\cite{rombach2022StableDiffusion} to a multi-view generative system, when pixel correspondences across views are available (e.g., generating perspective images to form a panorama). MVDream~\cite{shi2023mvdream} and Wonder3D~\cite{long2023wonder3d} further extend to more general settings where generated images yield 3D reconstruction via techniques such as NeRF~\cite{mildenhall2020nerf} or NeuS~\cite{wang2021neus}.

This paper pushes the frontier of multi-view diffusion models towards flexible and high-fidelity 3D reconstruction systems.
Concretely, the paper presents \ourmethod,
a novel approach to generate dense (32) and high-resolution (512$\times$512) images of an object, conditioned with single or sparse input views without camera poses, whose reliable estimation is difficult due to minimal or no visual overlaps. Standard 3D reconstruction techniques turn generated images into a 3D model. Two simple ideas are at the heart of our method. First, we leverage a latent diffusion inpainting model with conditional and generation branches, where self-attention among 2D features learns 3D consistency without using camera poses or image projection formula. Second, we introduce ``view dropout'' training strategy, which randomly excludes generation views in each batch, enabling the use of high-resolution images during training. During testing, this simple approach surprisingly generates high-quality, dense views for all the images simultaneously.

\ourmethod~achieves state-of-the-art performance on the task of novel view synthesis, single-view reconstruction, and sparse-view reconstruction. For single-view reconstruction, our method achieves 0.6973 IoU and 0.0165 Chamfer distance on the Google Scanned Objects dataset, higher than SyncDreamer~\cite{liu2023syncdreamer} by 0.1552 in terms of Vol. IOU. For novel view synthesis in sparse view setting, \ourmethod~improves the PSNR by 8.19 compared with a recent pose-free view synthesis method, LEAP~\cite{jiang2023leap}. 
Lastly, we demonstrate applications in text-to-3D by combining {\ourmethod} with a text-to-image generative model.

\section{Related work}
\label{sec:related_work}

This paper presents a multi-view image generative model for object reconstruction, given one or a few condition images. The section reviews related work on multi-view image generation and single to sparse-view 3D reconstruction techniques.

\mypara{Multi-view image generation}
The evolution of text-to-image diffusion models has paved the way for multi-view image generation. MVDiffusion~\cite{tang2023mvdiffusion} introduces an innovative multi-branch Unet architecture for denoising multi-view images simultaneously. This approach, however, is constrained to cases with one-to-one image correspondences. Syncdreamer~\cite{liu2023syncdreamer} uses 3D volumes and depth-wise attention for maintaining multi-view consistency. 
MVDream~\cite{shi2023mvdream} takes a different path, incorporating 3D self-attention to extend the work to more general cases. Similarly, Wonder3D~\cite{long2023wonder3d} and Zero123++~\cite{shi2023zero123++} apply 3D self-attention to single-image conditioned multi-view image generation. These methods, while innovative, tend to produce sparse, low-resolution images due to the computational intensity of the attention mechanism. In contrast, our framework represents a more versatile solution capable of generating dense, high-resolution multi-view images conditioned on an arbitrary number of images.

\mypara{Single view reconstruction}
Single View Image Reconstruction is an active research area~\cite{liu2023syncdreamer,yan2016perspective,long2023wonder3d,mittal2022autosdf,wang2023slice3d, xu2023dmv3d}, driven by the advancements of generative models~\cite{liu2023syncdreamer, long2023wonder3d, wang2023slice3d, mittal2022autosdf}. Large reconstruction model~\cite{hong2023lrm} and DMV3D~\cite{xu2023dmv3d} predict triplanes from a single image, but the 3D volume limits its resolutions. 
%
The other method, Syncdreamer~\cite{liu2023syncdreamer} generates multi-view images with a latent diffusion model by constructing a cost volume. These images are then used to recover 3D structures using conventional reconstruction methods like Neus. However, this process requires substantial GPU memory, limiting it to low resolutions. Similarly, Wonder3D faces challenges due to the computational demands of self-attention, leading to similar restrictions. In contrast, our approach introduces a "view dropout" technique, which randomly samples a limited number of views for training in each iteration. This enables our model to generate a variable number of high-resolution images while employing full 3D self-attention, effectively addressing the limitations faced by existing methods.

\mypara{Sparse view reconstruction}
Sparse View Image Reconstruction (SVIR)~\cite{jiang2023leap, yang2022fvor} is a challenging task where only a limited number of images, typically two to ten, are given. Traditional 3D reconstruction methods estimate camera poses first, then perform dense reconstruction using techniques such as multi-view stereo~\cite{stereopsis2010accurate, yao2018mvsnet} or NeRF~\cite{wang2021neus}.
However, camera pose estimation is difficult for SVIR, where visual overlaps are none to minimal. To address this, FvOR~\cite{yang2022fvor} optimizes camera poses and shapes jointly. LEAP~\cite{jiang2023leap} along with PF-LRM~\cite{wang2023pf-lrm} highlight the issues of noisy camera poses and suggest a pose-free approach. 
However, they are not based on generative models, lacking generative priors, and suffer from low-resolution outputs due to the use of volume rendering.
%
In contrast, our method 
employs a diffusion model to generate high-resolution multi-view images directly, then a reconstruction system Neus~\cite{wang2021neus} to recover a mesh model.

\section{Preliminary: Multi-view latent diffusion models}
\label{sec:preliminary}

MVDiffusion~\cite{tang2023mvdiffusion} is a multi-view latent diffusion model~\cite{tang2023mvdiffusion,shi2023mvdream,liu2023syncdreamer,shi2023zero123++}, generating multiple images given a text or an image, when pixel-wise correspondences are available across views.
MVDiffusion is the foundation of the proposed approach, where the section reviews its architecture and introduces notations (See \autoref{fig:equations}).

For generating eight perspective views forming a panorama, 
eight latent diffusion models (LDM) denoise eight noisy latent images $\{\latent_1(t), \latent_2(t), \cdots \latent_8(t) \}$ simultaneously.
%
A UNet is the core of a LDM model, 
consisting of a sequence of blocks through the four levels of the feature pyramid.

Let $U^i_b$ denote the feature image of $i$-th image at $b$-th block. A CNN initializes an input $U^0_i$ from $\latent_i(t)$ at the first block.
%
Each UNet block has four network modules.
The first is a novel correspondence-aware attention (CAA), enforcing consistency across views with visual overlaps: The left/right neighboring images ($U^b_{i-1}, U^b_{i+1}$) for panorama.
The remaining three modules are from the original:
1) Self-attention (SA) layers; 2) Cross-attention (CA) layers from the condition with the CLIP embedding; and 3) CNN layers with the pixel-wise concatenation of a positional encoding of time $\tau(t)$. At test time, a standard DDPM sampler~\cite{ho2020denoising} updates all noisy latents with the predicted noise from the last CNN layer.
The training objective is defined as follows by omitting the conditions for notation simplicity, where $\bepsilon^i$ is a Gaussian and $\model$ denotes the UNet output.
\begin{align}
\lsimplemv
:= \expec_{\left\{ \latent_i(0) \right\}_{i=1}^N, \left\{\bepsilon^i \sim \mathcal{N}(0, I)\right\}_{i=1}^N, t} \Big[ \sum_{i=1}^N \Vert \bepsilon^i - \model^i(\left\{ \latent_i(t) \right\}, \tau(t)) \Vert_{2}^{2}\Big].
\label{eq:cond_mvldm_loss}
\end{align}

\section{\ourmethod}
%

\ourmethod~pushes the frontier of multi-view diffusion models for 3D modeling in their \textit{flexibility} and \textit{scalability} by generating denser and higher-resolution images given an arbitrary number of un-posed condition views.
With the prevalence of Transformer models~\cite{vaswani2017transformer}, high-fidelity 3D modeling would require large-scale attention over dense and high-resolution image features, potentially with volumetric ones. Furthermore, 3D consistency learning is at the heart of the task, which would usually require precise image projection models and/or camera parameters. 
%
%
%
Our surprising discovery is that self-attention among 2D latent image features is all we need for 3D learning without projection models or camera parameters,
and a simple training strategy would further achieve dense and high-resolution multi-view image generation.
%
%
%
The section defines the task (i.e., input condition and output target images), then explains the two key ideas: 1) pose-free multi-view conditional diffusion model for flexibility and 2) view dropout training strategy for scalability.
\S\ref{sec:details} provides the remaining system details.

%

\subsection{Task: Input condition images and output target images}
\label{subsec:camera}

\begin{figure}[t]
    \centering
    \includegraphics[width=\textwidth]{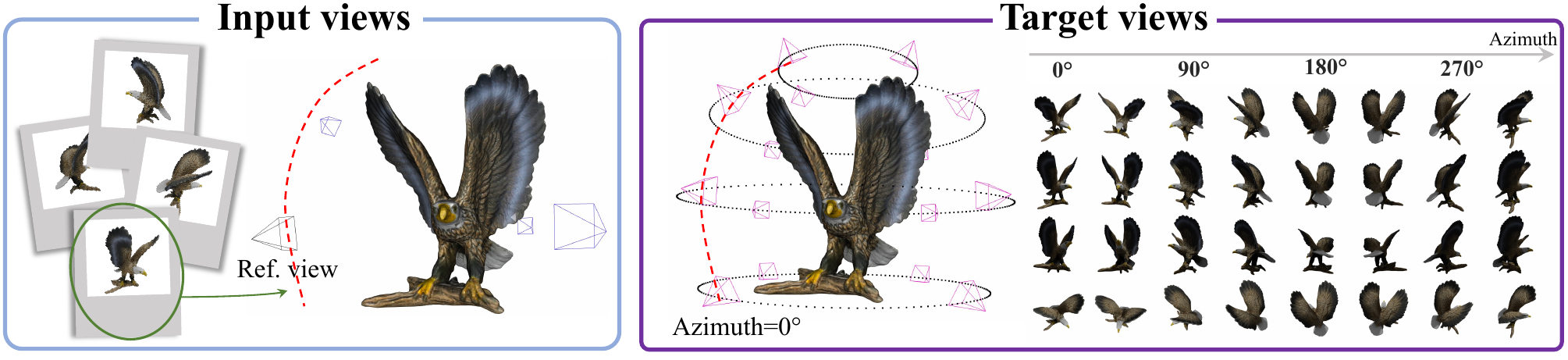}
    \caption{The input and output specification (\S\ref{subsec:camera}) of \ourmethod.
    The 32 target images are defined in eight azimuths and four elevation levels. During training, our view dropout strategy (\S\ref{subsec:view-drop}) randomly drops a substantial number of views (dashed blue) and trains the model to denoise the remaining views (red).
    }
    \label{fig:camera_configuration}
\end{figure}
\begin{figure}[t]
\centering
    \includegraphics[width=\textwidth]
    {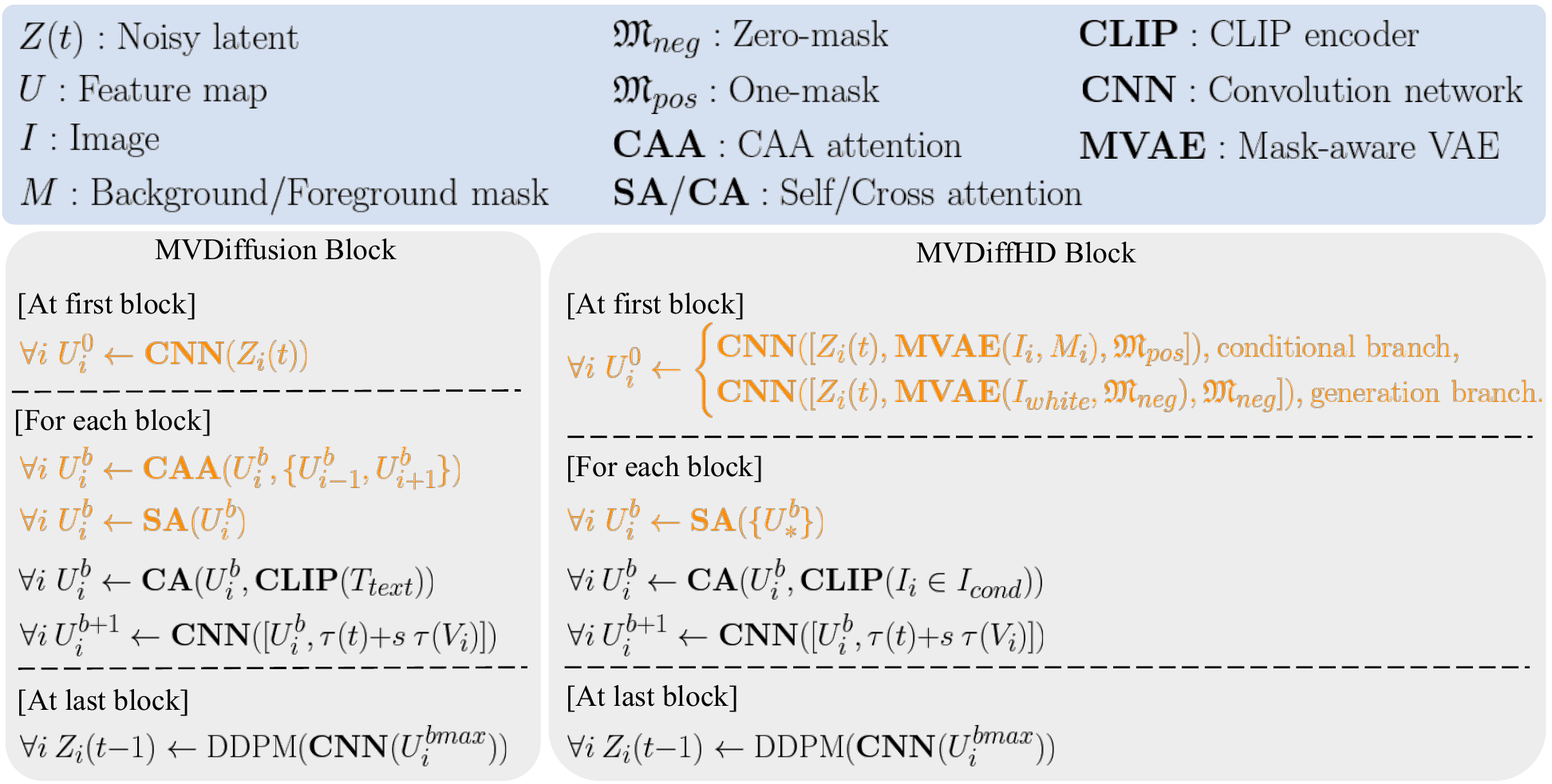}
    \caption{The denoising architectures for MVDiffusion and \ourmethod for sampling multi-view images. The order of the MVDiffusion network modules is rearranged to highlight the differences (in \textcolor{orange}{orange}) with \ourmethod.
    }
    \label{fig:equations}
\end{figure}

%

\begin{figure}[t]
\centering
\includegraphics[width=\textwidth]{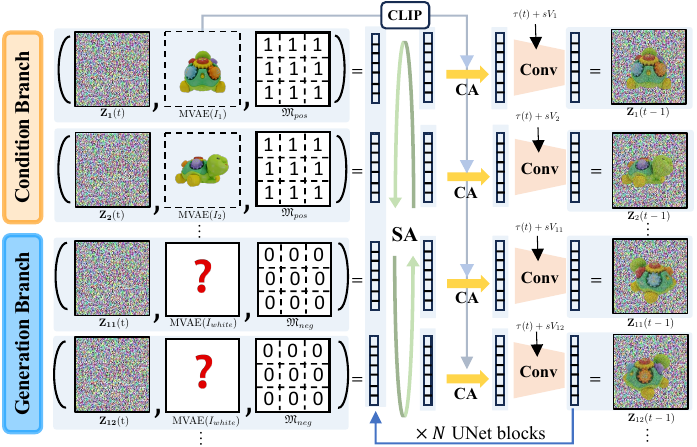}
\caption{Illustration of the pose-free multi-view conditional diffusion model of \ourmethod. The model takes any number of input images and generates images at fixed viewpoints. The condition branch and generation branch have different input configurations but share the same structure and weights. 
}
\label{fig:overview}
\end{figure}


%

The generation target is a set of dense (32) and high-resolution (512$\times$512) images, positioned at uniform 2D grid points on a sphere along the azimuth and elevation angles for 3D object reconstruction (See \autoref{fig:camera_configuration}). Specifically, there are eight azimuth angles (every 45$^\circ$) and four elevation angles (every 30$^\circ$ in the range $[-30^\circ, 60^\circ]$). 
Camera up-vectors are aligned with gravity, and their optical axes pass through the sphere center.
Our input condition is one or a few images without camera poses, where visual overlaps are too minimal or possibly none for Structure from Motion algorithms to work reliably. The number of condition images is up to a pre-determined number, which is $10$ in our experiments but can easily change. The input image resolution is 512$\times$512. The horizontal and vertical field-of-view of both the input and output views is 60$^\circ$.


We use synthetic rendered images from 3D object databases for training and evaluations, then real-world images as the input condition for further qualitative evaluations.
The task settings vary slightly between datasets, with details provided in \S\ref{sec:details}.
Here, we explain one preprocessing step that removes ambiguity in the training task.
3D object databases such as Objaverse~\cite{deitke2023objaverse} and Google Scanned Object~\cite{downs2022gso} align the Z-axis with the object up-vectors. 
However, the azimuth of the ground-truth object pose is ambiguous without 
camera poses of the condition images.
Therefore, we rotate the output views to align the azimuth of the first condition and the first output image as shown by the right of \autoref{fig:camera_configuration}.



%


\subsection{Pose-free multi-view conditional diffusion model}
\label{subsec:dm}




\ourmethod~is a multi-view latent diffusion model as defined in \S\ref{sec:preliminary}, comprising of a \textit{condition branch} for single or sparse-view input images and a \textit{generation branch} for output images (See \autoref{fig:equations} and \autoref{fig:overview}). Note that the condition branch shares the same architecture and is tasked to generate the condition images that are also given as guidance (i.e., a trivial task).

\mypara{Diffusion process}
The forward diffusion process is the same as MVDiffusion, except for the image resolution and the pre-trained VAE. Concretely, it 1) converts all $512\times 512\times 3$ input/output image ($I_i$) with foreground masks to $64\times 64\times 4$ latent images ($Z_i$) by a fine-tuned latent diffusion VAE (denoted as MVAE, see \S\ref{sec:details} for the fine-tuning process); and 2) adds a Gaussian noise with a linear schedule, as suggested by zero-123++~\cite{shi2023zero123++} to each feature of $Z_i$.

\mypara{Denoising process}
The denoising process is highlighted in \autoref{fig:equations}, where a latent diffusion UNet with a few modifications processes a noisy latent $Z_i(t)$ at each denoising step $t$. The UNet consists of $9$ blocks of network modules over the four levels of feature pyramids on either side of the encoder/decoder. The details are explained as follows.

\myparapara{[At first block]}
The UNet feature $U^0_i$ at the first block is initialized with the concatenation of 1) the noisy latents $Z_i(t)$; 2) a constant binary mask of either 1 or 0, denoted by $\mathfrak{M}_{pos}$ or $\mathfrak{M}_{neg}$ to indicate the branch type (condition or generation); and 3) the condition latents ($\text{MVAE}(I_i, M_i)$) where we use the conditonal VAE from latent diffusion to encode the condition image ($I_i$) with its segmentation mask ($M_i$). 
Note that this concatenation has $9=(4+4+1)$ channels, and a $1\times1$ final convolution layer reduces the channel dimension to $4$.
For a generation branch, we pass a white image 
as $I_i$ and a binary image of 1 (i.e., $\mathfrak{M}_{pos}$) as $M_i$. 
For Objaverse and Google Scaned Object datasets, we use the masks provided by the datasets. Otherwise, we run segmentation to generate the masks.

\myparapara{[For each block]}
Three network modules are at the heart of the processing: 1) Global self-attention mechanism among the UNet features across all the images, learning 3D consistency; 2) Cross-attention mechanism, injecting the CLIP embedding of the condition images to all the other images through the CLIP embedding; and 3) CNN layers, process per-image features while injecting the timestep frequency encoding $\tau(t)$ and the learnable embedding of an image index $V_i$.
%
For the self-attention module, we copy the network architecture and model weights 
and apply it across all the views. This module is inspired by MVDream~\cite{shi2023mvdream}, while the key differences in our work are 1) Scalability deployment via the view-drop training strategy in \S\ref{subsec:view-drop}; and 2) Handling of multiple condition images without camera poses via the network design.
%
%
%
$42=(32+10)$ learnable embedding vectors $\{V_{i}\}$ are trained for 32 generation and 10 condition images, each of which is multiplied with a zero-initialized trainable scale $s$ to avoid model disruption at the start of training.
%

\myparapara{[At last block]} The output of the last UNet block yields the noise estimation, and a standard DDPM sampler~\cite{ho2020denoising} takes it to produce the noisy latent of the next timestep $Z_i(t-1)$ for each sampling step. The loss function is the same as MVDiffusion. Note that the model is first trained with $\epsilon$-prediction and then with v-prediction (See \S\ref{sec:details}), where \autoref{eq:cond_mvldm_loss} is the loss function for the $\epsilon$-prediction model. The velocity~\cite{salimans2022progressive-v-pred}, $\bv^i(t) = \alpha_t \bepsilon^i - \gamma_t \latent_i(0) $,  becomes the prediction target for the v-prediction model, while $\alpha_t$ and $\gamma_t$ are predefined angular parameters.


\subsection{View dropout training strategy}
\label{subsec:view-drop}
\ourmethod~training would face a scalability challenge.
%
$42(=32+10)$ copies of UNet features yield more than 130k tokens, where the global self-attention mechanism becomes infeasible even with the latest memory efficient transformers for large language models~\cite{dao2022flashattention,dao2023flashattention2}.
We propose a simple yet surprisingly effective \textit{view dropout} training strategy, which
completely discards a set of views across all layers during training. Specifically, we randomly drop 24 out of 32 views for each object at each training iteration, significantly reducing memory consumption at training.
At test time, we run the entire architecture and generate 32 views.

\section{Remaining system details}
\label{sec:details}
This section explains the remaining system details on the data preparations, the mesh extraction process, the MVAE pre-fine-tuning, and the three-stage training strategy.

\subsection{Training data preparation}
Out of 800k 3D object models from Objaverse~\cite{deitke2023objaverse}, we use 180k models whose aesthetic scores~\cite{ava} are at least 5 for training.
%
For each object 3D model, we translate the bounding box center to the origin and apply uniform scaling so that the longest dimension matches $[-1, 1]$. The output camera centers are placed at a distance of 1.5 from the origin.
Input condition views are chosen in a similar way as Zero-123~\cite{liu2023zero}. 
Concretely, an azimuth angle is randomly chosen from one of the eight discrete angles of the output cameras (also see \S\ref{subsec:camera}). The elevation angle is set randomly from [-10$^\circ$, 45$^\circ$]. The distance of the camera center from the origin is set randomly from [1.5, 2.2]. We use Blender to render images. 



\subsection{Testing data preparation}
\mypara{Single-view cases}
Google Scanned Object (GSO)~\cite{downs2022gso} is our testing dataset, where we borrow the rendered images and the evaluation pipeline
from SyncDreamer~\cite{liu2023syncdreamer}. Concretely, the test set consists of 30 objects. Each object has 16 images with a fixed elevation of 30$^\circ$ and every 22.5$^\circ$ for azimuth. SyncDreamer selected condition images by ``visual plausibility", which we copy. Since the azimuth angles in our training setting are every 45$^\circ$, 
eight images (starting from and including the condition image) are used for evaluation. The resolution of the rendered images is 256x256, while the image resolution of our architecture is 512x512. We upscale the condition images to 512x512 for our system inputs. The ground-truth images are 256x256 and we downscale our generated images to 256x256 for evaluation, while 512x512 images are used for the mesh reconstruction.
%
The Chamfer Distances (CD) and volume IoU between the ground-truth and reconstructed shapes are reported for single-view 3D reconstruction. The PSNR, SSIM~\cite{wang2004ssim}, and LPIPS~\cite{zhang2018lpips} are reported for novel view synthesis (NVS) 
by averaging over the eight images.

\mypara{Sparse-view cases}
Sparse-view un-posed condition is a new setup (except the work of LEAP~\cite{jiang2023leap} and PF-LRM~\cite{wang2023pf-lrm} to our knowledge). We use a process similar to the single-view setting to render images.
Concretely, we first 
render 10 condition images for each of the 30 GSO objects. The azimuth and the elevation angles are chosen randomly from [0, 360) and [-10, 45] respectively. 
We render 32 ground-truth target images while aligning the azimuth of the first target view and the first input view (See \S\ref{subsec:camera} and Fig.~\ref{fig:camera_configuration}). The same evaluation metrics are used, while we vary the number of condition images to be 1, 2, 4, 8, and 10.


\subsection{Mesh extraction from generated images}
After generating 32 images (all target views in Fig.~\ref{fig:camera_configuration}), a neural implicit reconstruction method recovers a mesh model, similar to SyncDreamer~\cite{liu2023syncdreamer} and Wonder3D~\cite{long2023wonder3d}.
Specifically, we use grid-based NeuS~\cite{instant-nsr-pl,li2023neuralangelo}, where the foreground masks are decoded from the latent images $\{Z_i(0)\}$ by MVAE.
Since our generated images have high resolution and quality, we directly run the monocular normal estimator released by Omnidata~\cite{eftekhar2021omnidata} to obtain additional normal supervisions for NeuS without a normal generation module like Wonder3D. We borrow the NeuS implementation from Wonder3D's official codebase but do not use their ranking-based loss.
With a single Nvidia 2080 Ti, it takes around 3 minutes to reconstruct a textured mesh model.
The mesh could directly use the exported vertex color or be re-textured with the generated images.

\subsection{Mask-aware VAE pre-fine-tuning}
We copy the network architecture and model weights of the default VAE~\cite{rombach2022high} and add additional input and output channels to handle the mask.
We found that fine-tuning Mask-aware VAE (M-VAE) only with object images improves performance.
%
%
Concretely, we use approximately 3 million RGBA images rendered from Objaverse to fine-tune M-VAE as a pre-processing. We follow the original VAE hyperparameters with a base learning rate of 4.5e-6 and a batch size of 64. The training runs for 60,000 iterations.
%
The binary cross entropy loss is used for the mask channel. The process improves PSNR from 36.6 to 41.2.

\subsection{Three-stage training strategy} After initializing the UNet model weights by a pre-trained latent diffusion inpainting model, we train the proposed system in three stages. 
First, we train as an $\epsilon$-prediction model only with single-view conditioning cases, because our pre-trained model was trained as $\epsilon$-prediction.
Second, we fine-tune as a v-prediction model~\cite{salimans2022progressive-v-pred} still with single-view conditioning cases. Third, we fine-tune as a v-prediction model with both single and sparse-view conditioning cases.
Half the samples are single-view conditioning, and the other half are sparse-view conditioning, where the number of condition images is uniformly sampled between 2 and 10.

\section{Experiments}\label{exp_sec}


We train the model with a batch size of 1024 using 128 Nvidia H100 GPUs for about a week. 
At test time, we use DDPM~\cite{ho2020denoising} sampler with 75 steps to sample the multi-view images, and it takes our model 30s, 77s, 123s, and 181s to generate 8, 16, 24, and 32 images, respectively.  
The section presents 
the single view experiments in \S\ref{subsec:exp-single}, the sparse view experiments in \S\ref{subsec:exp-sparse}, and text-to-3D application experiments
in \S\ref{subsec:text_to_3d}.



\subsection{Single-view object modeling}
\label{subsec:exp-single}
\begin{table}[t]
    \centering
    \caption{
Single-view object modeling results, evaluating
reconstructed meshes (left) and generated images (right). The ground-truth meshes and images are prepared by SyncDreamer~\cite{liu2023syncdreamer} based on the Google Scanned Object~\cite{downs2022gso} dataset.
ICP is necessary to align reconstructed meshes for methods marked with $^*.$
}
    \label{tab:single-view-sync}
    \setlength{\tabcolsep}{4pt}
    \begin{tabular}{l@{\quad}cc@{\quad}ccc}
       \toprule
       \multicolumn{1}{l}{Task $\rightarrow$} &
\multicolumn{2}{c}{3D reconstruction} & \multicolumn{3}{c}{Novel view synthesis} \\
        \cmidrule(r){2-3}\cmidrule(r){4-6}
       Method  & Chamfer Dist.$\downarrow$ & Vol. IoU$\uparrow$ & PSNR$\uparrow$ & SSIM$\uparrow$ & LPIPS$\downarrow$  \\
       \midrule
       Realfusion~\cite{melas2023realfusion}    
       & 0.0819  & 0.2741 & 15.26 & 0.722 & 0.283  \\
       Magic123~\cite{qian2023magic123}
       & 0.0516 &  0.4528 & - & - & - \\
       One-2-3-45~\cite{liu2023one}    
       & 0.0629 &  0.4086 & - & - & - \\
       Point-E~\cite{nichol2022point}    
       & 0.0426 & 0.2875 & - & - & - \\
       Shap-E~\cite{jun2023shap}    
       & 0.0436 &  0.3584 & - & - & -  \\
       Zero123~\cite{liu2023zero}    
       & 0.0339 &  0.5035 & 18.93 & 0.779 & 0.166 \\
       SyncDreamer~\cite{liu2023syncdreamer}    
       &  {0.0261}  &  {0.5421} & {20.05} & {0.798} & {0.146}   \\
       Wonder3D~\cite{long2023wonder3d}$^*$ & 0.0329 & 0.5768 & - & - & - \\ 
       Open-LRM~\cite{openlrm}$^*$ & 0.0285 & 0.5945 & - & - & - \\
           Ours &  \textbf{0.0165}  &  \textbf{0.6973} &  \textbf{21.45} & \textbf{0.844} & \textbf{0.129} \\
       \bottomrule
    \end{tabular}
\end{table}
\begin{figure}[t]
\centering
\includegraphics[width=\textwidth]{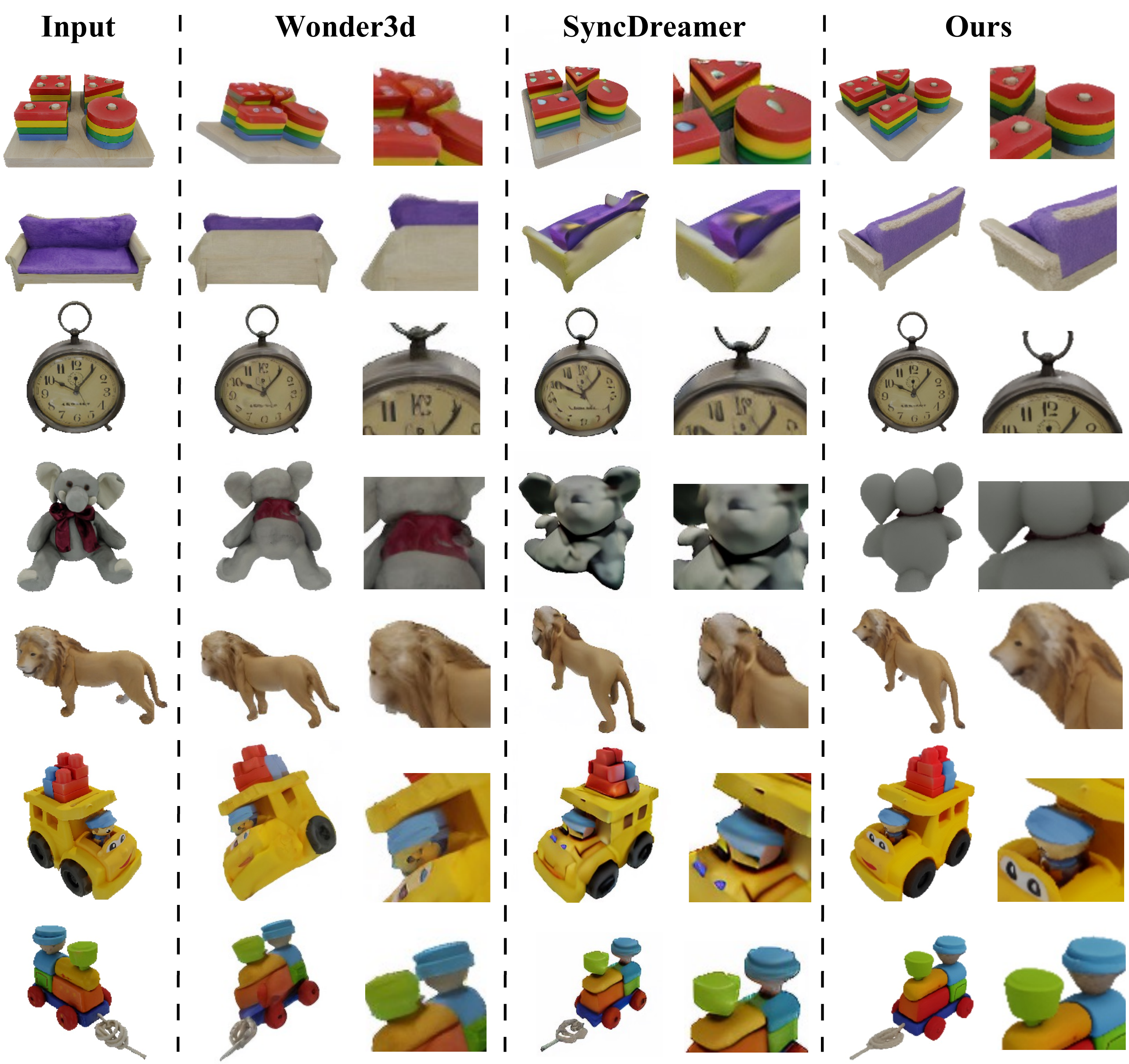}
\caption{Single-view object modeling results of generated images. The input image and the generated images by Wonder3D and SyncDreamer are in 256$\times$256. Our rendered images are in 512$\times$512, showing higher fidelity and richer details.
}
\label{fig:single-images-qualitative}
\end{figure}
%
Three state-of-the-art single-view object modeling methods are our main baselines: SyncDreamer~\cite{liu2023syncdreamer}, Wonder3D~\cite{long2023wonder3d}, and Open-LRM~\cite{openlrm}. 
Since the evaluation pipeline is the same as SyncDreamer, we copy numbers of other baselines in their paper for comparison,
which includes Zero123~\cite{liu2023zero}, RealFusion~\cite{melas2023realfusion}, Magic123~\cite{qian2023magic123}, One-2-3-45~\cite{liu2023one}, Point-E~\cite{nichol2022point}, and Shap-E~\cite{jun2023shap}. The following introduces the three main baselines and how we reproduce their systems:

\noindent \raisebox{0.25ex}{\tiny$\bullet$} {\it SyncDreamer} generates 16 images from fixed viewpoints given a single input image. The image resolution is 256x256. Their denoising network $\model$ initializes from Zero123 and leverages 3D feature volumes and depth-wise attention to learn multi-view consistency. 
It requires users to provide the elevation of the input image. 

\noindent \raisebox{0.25ex}{\tiny$\bullet$} {\it Wonder3D} takes a single input image as the canonical view and generates 6 images as well as the normal maps. The image resolution is 256$\times$256. Multi-view self-attention and an extra cross-domain attention ensure the consistency of generation results, while the views are sparser than ours.  We run the official codebase on the GSO input images to get the results. However, the released model assumes orthographic cameras and we cannot use the same test set to evaluate the NVS performance. ICP aligns the reconstructed mesh with the ground truth before computing the metrics.  

\noindent \raisebox{0.25ex}{\tiny$\bullet$} {\it Open-LRM} is an open-source implementation of Large Reconstruction Model (LRM)~\cite{hong2023lrm}, a generalized reconstruction model that predicts a triplane NeRF from a single input image using a feed-forward transformer-based network. ICP aligns the reconstructed mesh with the ground truth before computing the CD and volume IoU.

\begin{figure}[t]
\centering
\includegraphics[width=\textwidth]{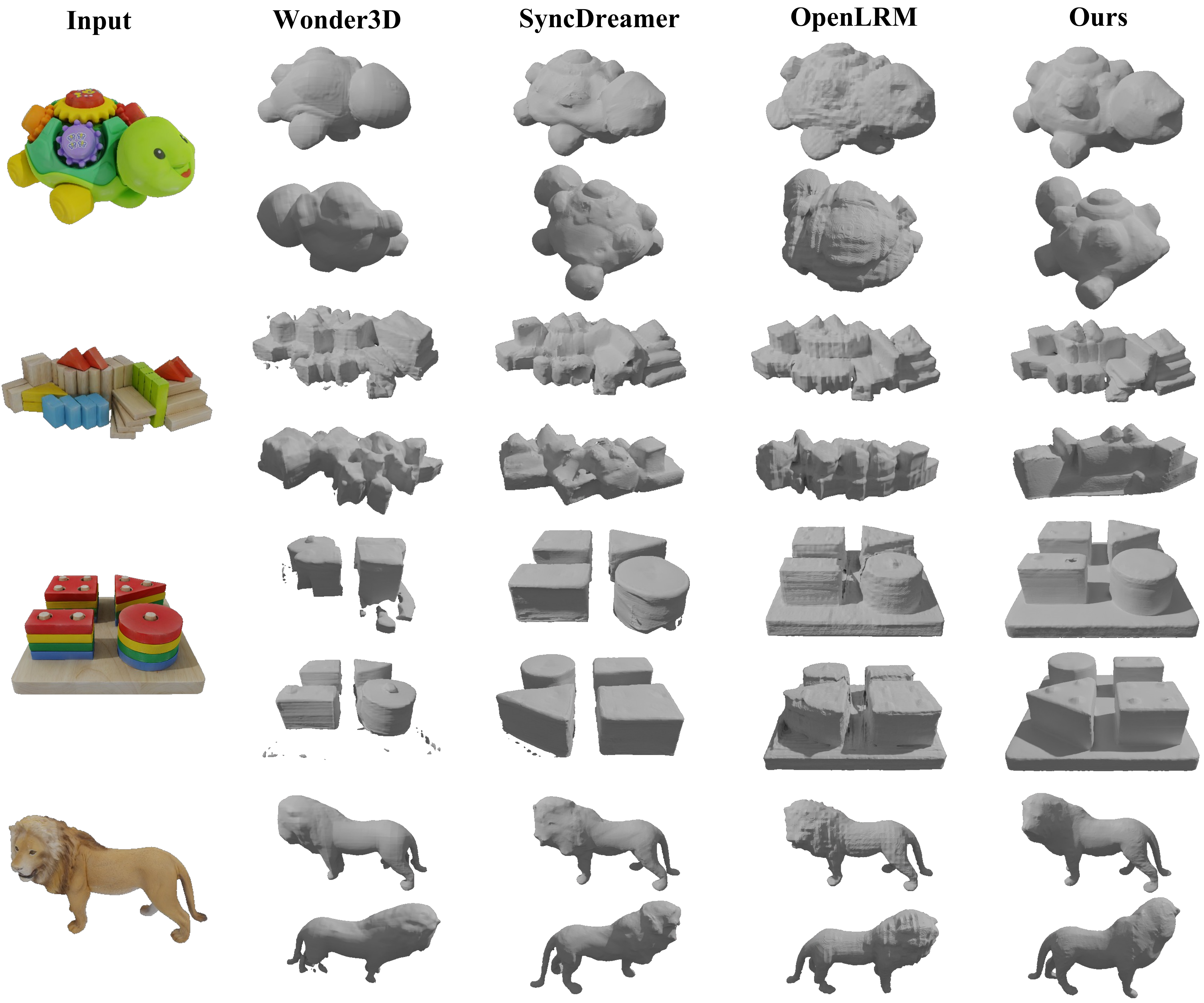}
\caption{
%
Single-view object modeling results of reconstructed mesh models. Our meshes are exported from dense (32) and high-resolution (512$\times$512) generated images, demonstrating finer details.
%
}
\label{fig:single-mesh-qualitative}
\end{figure}

\mypara{Results} \autoref{tab:single-view-sync} presents the quantitative evaluations of the reconstructed 3D meshes and the generated images.
\ourmethod~consistently outperforms all the competing methods with clear margins.
Note that the evaluation is not completely fair for Wonder3D that assumes orthographic camera projections, where perspective images are used in the experiments. However, 
we believe the clear performance gaps suffice to demonstrate the strength of our method.

\autoref{fig:single-images-qualitative} and \autoref{fig:single-mesh-qualitative} show generated images and reconstructed mesh models. In \autoref{fig:single-images-qualitative}, our method clearly shows the number on the clock (row 3), while others exhibit blurry numbers. Another example (row 5) showcases two perfectly symmetrical windows generated by our method, contrasting with Wonder3D's failure to maintain symmetry or clarity. In \autoref{fig:single-mesh-qualitative}, our method can recover a plausible and detailed shape of the turtle example (row 1), while Wonder3D and OpenLRM fail to recognize it as a turtle and exhibit significant artifacts. 
%

\subsection{Sparse-view object modeling}
\label{subsec:exp-sparse}

\begin{table}[tbh]
\caption{
Sparse-view object modeling results, evaluating
reconstructed meshes (left) and generated images (right), based on the GSO~\cite{downs2022gso} dataset.
}
\label{tab:sparse-view-quantitative}
\begin{minipage}{0.51\textwidth}
    \centering
    \setlength{\tabcolsep}{2.5pt}
    \resizebox{\linewidth}{!}{
    \begin{tabular}{cc@{\quad}cc}
       \toprule
       Method & Views & Chamfer Dist.$\downarrow$ & Vol. IoU$\uparrow$  \\
       \midrule
       \multirow{2}*{\shortstack[c]{Sync-\\ Dreamer}} & \multirow{2}*{1} & \multirow{2}*{0.0318} & \multirow{2}*{0.5610} \\
       \\
       \midrule
       \multirow{6}*{\shortstack[c]{NeuS\cite{wang2021neus}\\ (G.T. pose)}} & 1 & 0.0536 & 0.4400  \\
       & 2 & 0.0307 & 0.5884 \\
       & 4 & 0.0158 & 0.7323 \\
       & 10 & 0.0096 & 0.8092 \\ 
       \midrule
       \multirow{6}*{Ours} & 1 & 0.0208 & 0.6689 \\
       & 2 & 0.0158 & 0.7260 \\
       & 4 & 0.0122 & 0.7737  \\
       & 10 & 0.0101  & 0.8046    \\
       \bottomrule
    \end{tabular}
    }
\end{minipage}
\begin{minipage}{0.48\textwidth}
    \centering
    \setlength{\tabcolsep}{2.5pt}
    \resizebox{\linewidth}{!}{
    \begin{tabular}{c@{\quad}c@{\quad}ccc}
       \toprule
       Method & Views & PSNR$\uparrow$ & SSIM$\uparrow$ & LPIPS$\downarrow$  \\
       \midrule
       \multirow{2}*{\shortstack[c]{Sync-\\ Dreamer}} & \multirow{2}*{1} & \multirow{2}*{19.46} & \multirow{2}*{0.847} & \multirow{2}*{0.188} \\
       \\
       \midrule
       \multirow{6}*{LEAP\cite{jiang2023leap}}  & 1 & 14.66 & 0.47 &0.43\\
       & 2 &16.22 &0.59 &0.36 \\
       & 4 & 16.54&0.61&0.35 \\
       & 10 &16.84&0.64&0.34\\
       \midrule
       \multirow{6}*{Ours} & 1 & 20.25 & 0.862 & 0.157 \\
       & 2 & 21.73 & 0.872 & 0.137 \\
       & 4  & 23.44 & 0.886 & 0.117 \\
       & 10 & 25.03 & 0.899 & 0.102 \\
       
       \bottomrule
    \end{tabular}
    }
\end{minipage}
\end{table}
\begin{figure}[!t]
\centering
\includegraphics[width=\textwidth]{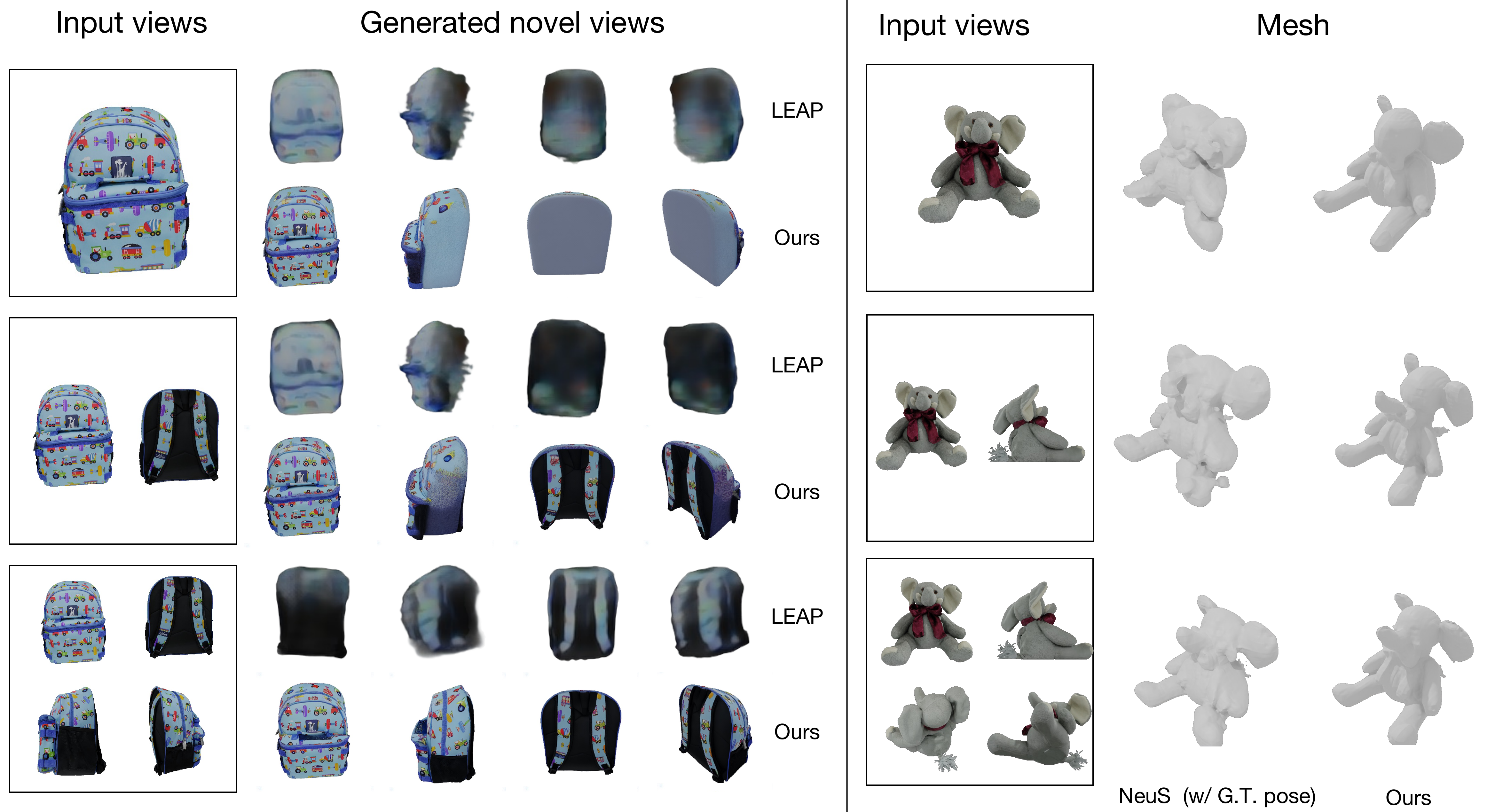}
\caption{Novel view synthesis and 3D reconstruction with sparse-view input images. \textbf{Left}: a qualitative example of novel view synthesis, comparing  LEAP~\cite{jiang2023leap} and \ourmethod~with different numbers of unposed input images. \textbf{Right}: qualitative comparison of reconstructed meshes between NeuS~\cite{wang2021neus} with ground-truth relative poses and our pose-free \ourmethod.}
\label{fig:sparse-qualitative}
\end{figure}
\begin{figure}[!t]
    \centering
    \includegraphics[width=0.9\textwidth]{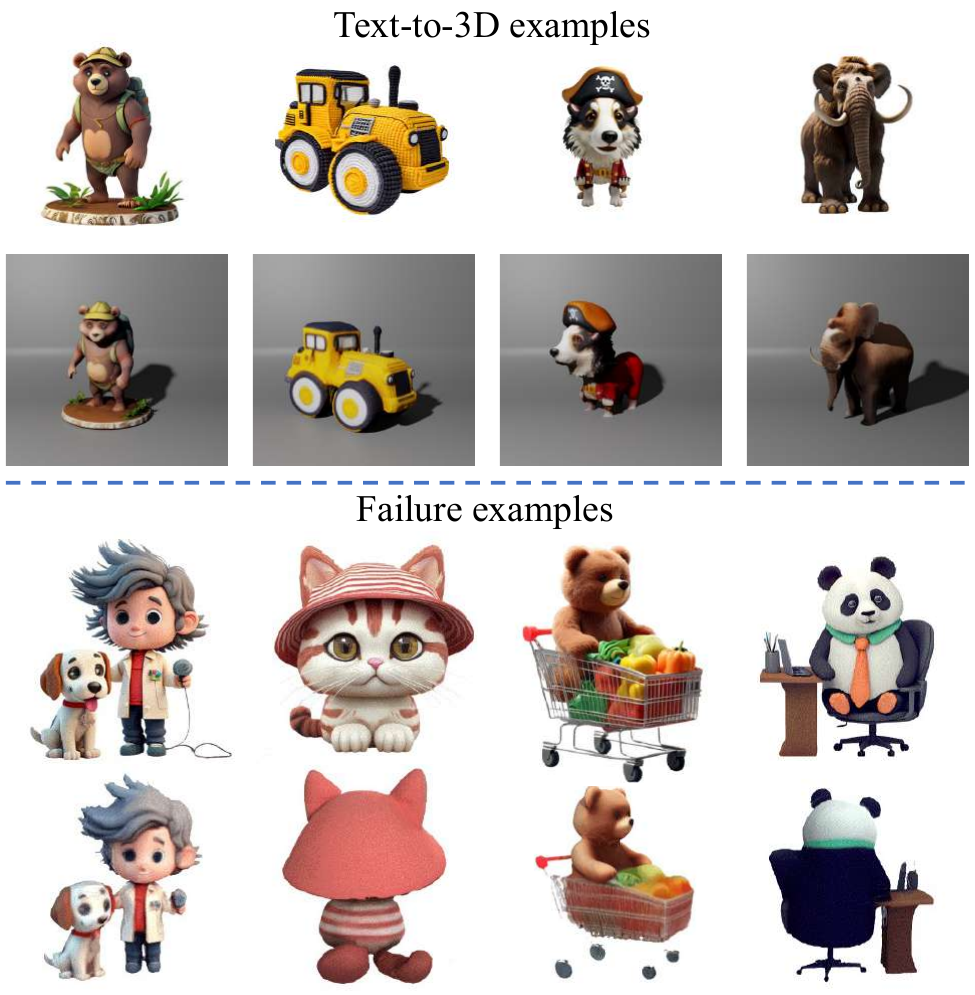}
    \caption{Text-to-3D application examples. (Top) A text-to-image model generates an image given a text-prompt. (Bottom) {\ourmethod} turns the generated image into a 3D model. We also show some failure examples in the bottom. }
    \label{fig:text_to_3d}
\end{figure}



Sparse-view un-posed input images is a challenging setting, where we are aware of only a few existing approaches such as LEAP~\cite{jiang2023leap} and PF-LRM~\cite{wang2023pf-lrm}, a sparse-view pose-free extension of LRM~\cite{hong2023lrm}. There is no public implementation of PF-LRM, and we pick LEAP as the first baseline. The literature on multi-view 3D reconstruction is extensive. It would be valuable to contrast our approach, even though they require camera poses as input. As a compromise, we have selected NeuS~\cite{wang2021neus} as our second benchmark by providing the ground-truth camera poses as their input.


%

\noindent \raisebox{0.25ex}{\tiny$\bullet$} {\it LEAP} leverages a transformer to predict neural volumes of radiance fields from a sparse number of views and is also pose-free. LEAP employs DINOv2~\cite{oquab2023dinov2} as the feature extractor and has reasonable generalization capacity.

\noindent \raisebox{0.25ex}{\tiny$\bullet$} {\it NeuS}
is a 3D reconstruction method, where we provide the ground-truth camera poses of the condition images as well as surface normals estimated by Omnidata's monocular normal estimator~\cite{eftekhar2021omnidata}. We use the public grid-based NeuS implementation~\cite{instant-nsr-pl}. 
This baseline is similar to MonoSDF~\cite{yu2022monosdf} or NeurIS~\cite{wang2022neuris} equipped with ground-truth foreground masks and camera poses, thus sets a performance upper bound for methods without generative priors.


\mypara{Results} \autoref{tab:sparse-view-quantitative} and \autoref{fig:sparse-qualitative} present the quantitative and qualitative comparison results, respectively. Compared to LEAP, \ourmethod~generates images with much better quality. LEAP and our method both exploit multi-view self-attention to establish global 3D consistency. Therefore, we attribute our better performance to the strong image priors inherited from the pre-trained latent diffusion models.
Our reconstructed meshes outperform NeuS in most settings, a notable achievement considering that NeuS uses ground-truth camera poses. This comparison highlights the practicality of our method, enabling users to achieve high-quality 3D models from just a few object snapshots.

%



\subsection{Text-to-3D application}
\label{subsec:text_to_3d}

\ourmethod~shows consistent performance with minimal errors on the GSO dataset. Note that our training data solely comes from Objaverse dataset~\cite{deitke2023objaverse}, and \ourmethod~already achieves remarkable generalization capabilities. 
To further challenge the system, we demonstrate a text-to-3D application, where a text-to-image model prepares an input condition image. \ourmethod~turns the condition image into a 3D model.
\autoref{fig:text_to_3d} has four examples demonstrating the power of our approach.


\section{Limitations and future challenges}
This paper presents a pose-free technique for reconstructing objects using an arbitrary number of images. Central to this approach is a sophisticated multi-branch, multi-view diffusion model. This model processes any number of conditional images to produce dense, consistent views from fixed perspectives. This capability significantly enhances the performance of existing reconstruction algorithms, enabling them to generate high-quality 3D models. Our results show that \ourmethod ~sets a new standard in performance for both single-view and sparse-view object reconstruction.

\autoref{fig:text_to_3d} presents typical failure modes and the limitations of our approach.
Our method struggles with thin structures
as in the leftmost example, which fails to reconstruct a cable.
Our method occasionally generates implausible images for views occluded in the input,
a notable instance being the depiction of a cat with two tails.
%
These shortcomings are predominantly attributed to the lack of
training data, where
one future work will expand the framework to incorporate videos, which offer richer contextual and spatial information, potentially enabling dynamic video generation.

\mypara{Acknowledgements} This research is partially supported by NSERC Discovery Grants with Accelerator Supplements and DND/NSERC Discovery Grant Supplement, NSERC Alliance Grants, and John R. Evans Leaders Fund (JELF). We thank the Digital Research Alliance of Canada and BC DRI Group for providing computational resources.


\clearpage
%
%

\bibliographystyle{splncs04}
\bibliography{egbib}

\begin{thebibliography}{10}
\providecommand{\url}[1]{\texttt{#1}}
\providecommand{\urlprefix}{URL }
\providecommand{\doi}[1]{https://doi.org/#1}

\bibitem{agarwal2011building}
Agarwal, S., Furukawa, Y., Snavely, N., Simon, I., Curless, B., Seitz, S.M.,
  Szeliski, R.: Building rome in a day. Communications of the ACM
  \textbf{54}(10),  105--112 (2011)

\bibitem{dao2023flashattention2}
Dao, T.: Flashattention-2: Faster attention with better parallelism and work
  partitioning. arXiv preprint arXiv:2307.08691  (2023)

\bibitem{dao2022flashattention}
Dao, T., Fu, D., Ermon, S., Rudra, A., R{\'e}, C.: Flashattention: Fast and
  memory-efficient exact attention with io-awareness. Advances in Neural
  Information Processing Systems  \textbf{35},  16344--16359 (2022)

\bibitem{deitke2023objaverse}
Deitke, M., Schwenk, D., Salvador, J., Weihs, L., Michel, O., VanderBilt, E.,
  Schmidt, L., Ehsani, K., Kembhavi, A., Farhadi, A.: Objaverse: A universe of
  annotated 3d objects. In: Proceedings of the IEEE/CVF Conference on Computer
  Vision and Pattern Recognition. pp. 13142--13153 (2023)

\bibitem{downs2022gso}
Downs, L., Francis, A., Koenig, N., Kinman, B., Hickman, R., Reymann, K.,
  McHugh, T.B., Vanhoucke, V.: Google scanned objects: A high-quality dataset
  of 3d scanned household items. In: ICRA (2022)

\bibitem{eftekhar2021omnidata}
Eftekhar, A., Sax, A., Malik, J., Zamir, A.: Omnidata: A scalable pipeline for
  making multi-task mid-level vision datasets from 3d scans. In: Proceedings of
  the IEEE/CVF International Conference on Computer Vision. pp. 10786--10796
  (2021)

\bibitem{furukawa2010towards}
Furukawa, Y., Curless, B., Seitz, S.M., Szeliski, R.: Towards internet-scale
  multi-view stereo. In: 2010 IEEE computer society conference on computer
  vision and pattern recognition. pp. 1434--1441. IEEE (2010)

\bibitem{instant-nsr-pl}
Guo, Y.C.: Instant neural surface reconstruction (2022),
  https://github.com/bennyguo/instant-nsr-pl

\bibitem{openlrm}
He, Z., Wang, T.: Openlrm: Open-source large reconstruction models.
  \url{https://github.com/3DTopia/OpenLRM} (2023)

\bibitem{ho2020denoising}
Ho, J., Jain, A., Abbeel, P.: Denoising diffusion probabilistic models.
  Advances in neural information processing systems  \textbf{33},  6840--6851
  (2020)

\bibitem{hong2023lrm}
Hong, Y., Zhang, K., Gu, J., Bi, S., Zhou, Y., Liu, D., Liu, F., Sunkavalli,
  K., Bui, T., Tan, H.: Lrm: Large reconstruction model for single image to 3d.
  arXiv preprint arXiv:2311.04400  (2023)

\bibitem{jiang2023leap}
Jiang, H., Jiang, Z., Zhao, Y., Huang, Q.: Leap: Liberate sparse-view 3d
  modeling from camera poses. arXiv preprint arXiv:2310.01410  (2023)

\bibitem{jun2023shap}
Jun, H., Nichol, A.: Shap-e: Generating conditional 3d implicit functions.
  arXiv preprint arXiv:2305.02463  (2023)

\bibitem{li2023neuralangelo}
Li, Z., M{\"u}ller, T., Evans, A., Taylor, R.H., Unberath, M., Liu, M.Y., Lin,
  C.H.: Neuralangelo: High-fidelity neural surface reconstruction. In:
  Proceedings of the IEEE/CVF Conference on Computer Vision and Pattern
  Recognition. pp. 8456--8465 (2023)

\bibitem{liu2023one2345++}
Liu, M., Shi, R., Chen, L., Zhang, Z., Xu, C., Wei, X., Chen, H., Zeng, C., Gu,
  J., Su, H.: One-2-3-45++: Fast single image to 3d objects with consistent
  multi-view generation and 3d diffusion. arXiv preprint arXiv:2311.07885
  (2023)

\bibitem{liu2023one}
Liu, M., Xu, C., Jin, H., Chen, L., Xu, Z., Su, H.: One-2-3-45: Any single
  image to 3d mesh in 45 seconds without per-shape optimization. arXiv preprint
  arXiv:2306.16928  (2023)

\bibitem{liu2023zero}
Liu, R., Wu, R., Van~Hoorick, B., Tokmakov, P., Zakharov, S., Vondrick, C.:
  Zero-1-to-3: Zero-shot one image to 3d object. In: ICCV (2023)

\bibitem{liu2023syncdreamer}
Liu, Y., Lin, C., Zeng, Z., Long, X., Liu, L., Komura, T., Wang, W.:
  Syncdreamer: Generating multiview-consistent images from a single-view image.
  arXiv preprint arXiv:2309.03453  (2023)

\bibitem{long2023wonder3d}
Long, X., Guo, Y.C., Lin, C., Liu, Y., Dou, Z., Liu, L., Ma, Y., Zhang, S.H.,
  Habermann, M., Theobalt, C., et~al.: Wonder3d: Single image to 3d using
  cross-domain diffusion. arXiv preprint arXiv:2310.15008  (2023)

\bibitem{melas2023realfusion}
Melas-Kyriazi, L., Laina, I., Rupprecht, C., Vedaldi, A.: Realfusion: 360deg
  reconstruction of any object from a single image. In: CVPR (2023)

\bibitem{mildenhall2020nerf}
Mildenhall, B., Srinivasan, P.P., Tancik, M., Barron, J.T., Ramamoorthi, R.,
  Ng, R.: Nerf: Representing scenes as neural radiance fields for view
  synthesis. In: ECCV (2020)

\bibitem{mittal2022autosdf}
Mittal, P., Cheng, Y.C., Singh, M., Tulsiani, S.: Autosdf: Shape priors for 3d
  completion, reconstruction and generation. In: Proceedings of the IEEE/CVF
  Conference on Computer Vision and Pattern Recognition. pp. 306--315 (2022)

\bibitem{ava}
Murray, N., Marchesotti, L., Perronnin, F.: Ava: A large-scale database for
  aesthetic visual analysis. In: 2012 IEEE conference on computer vision and
  pattern recognition. pp. 2408--2415. IEEE (2012)

\bibitem{nichol2022point}
Nichol, A., Jun, H., Dhariwal, P., Mishkin, P., Chen, M.: Point-e: A system for
  generating 3d point clouds from complex prompts. arXiv preprint
  arXiv:2212.08751  (2022)

\bibitem{oquab2023dinov2}
Oquab, M., Darcet, T., Moutakanni, T., Vo, H., Szafraniec, M., Khalidov, V.,
  Fernandez, P., Haziza, D., Massa, F., El-Nouby, A., et~al.: Dinov2: Learning
  robust visual features without supervision. arXiv preprint arXiv:2304.07193
  (2023)

\bibitem{qian2023magic123}
Qian, G., Mai, J., Hamdi, A., Ren, J., Siarohin, A., Li, B., Lee, H.Y.,
  Skorokhodov, I., Wonka, P., Tulyakov, S., et~al.: Magic123: One image to
  high-quality 3d object generation using both 2d and 3d diffusion priors.
  arXiv preprint arXiv:2306.17843  (2023)

\bibitem{rombach2022StableDiffusion}
Rombach, R., Blattmann, A., Lorenz, D., Esser, P., Ommer, B.: High-resolution
  image synthesis with latent diffusion models. In: Proceedings of the IEEE/CVF
  Conference on Computer Vision and Pattern Recognition. pp. 10684--10695
  (2022)

\bibitem{rombach2022high}
Rombach, R., Blattmann, A., Lorenz, D., Esser, P., Ommer, B.: High-resolution
  image synthesis with latent diffusion models. In: Proceedings of the IEEE/CVF
  conference on computer vision and pattern recognition. pp. 10684--10695
  (2022)

\bibitem{salimans2022progressive-v-pred}
Salimans, T., Ho, J.: Progressive distillation for fast sampling of diffusion
  models. arXiv preprint arXiv:2202.00512  (2022)

\bibitem{shi2023zero123++}
Shi, R., Chen, H., Zhang, Z., Liu, M., Xu, C., Wei, X., Chen, L., Zeng, C., Su,
  H.: Zero123++: a single image to consistent multi-view diffusion base model.
  arXiv preprint arXiv:2310.15110  (2023)

\bibitem{shi2023mvdream}
Shi, Y., Wang, P., Ye, J., Long, M., Li, K., Yang, X.: Mvdream: Multi-view
  diffusion for 3d generation. arXiv preprint arXiv:2308.16512  (2023)

\bibitem{stereopsis2010accurate}
Stereopsis, R.M.: Accurate, dense, and robust multiview stereopsis. IEEE
  TRANSACTIONS ON PATTERN ANALYSIS AND MACHINE INTELLIGENCE  \textbf{32}(8)
  (2010)

\bibitem{tang2023mvdiffusion}
Tang, S., Zhang, F., Chen, J., Wang, P., Furukawa, Y.: Mvdiffusion: Enabling
  holistic multi-view image generation with correspondence-aware diffusion.
  arXiv preprint arXiv:2307.01097  (2023)

\bibitem{vaswani2017transformer}
Vaswani, A., Shazeer, N., Parmar, N., Uszkoreit, J., Jones, L., Gomez, A.N.,
  Kaiser, {\L}., Polosukhin, I.: Attention is all you need. Advances in neural
  information processing systems  \textbf{30} (2017)

\bibitem{wang2022neuris}
Wang, J., Wang, P., Long, X., Theobalt, C., Komura, T., Liu, L., Wang, W.:
  Neuris: Neural reconstruction of indoor scenes using normal priors. In:
  European Conference on Computer Vision. pp. 139--155. Springer (2022)

\bibitem{wang2021neus}
Wang, P., Liu, L., Liu, Y., Theobalt, C., Komura, T., Wang, W.: Neus: Learning
  neural implicit surfaces by volume rendering for multi-view reconstruction.
  In: NeurIPS (2021)

\bibitem{wang2023imagedream}
Wang, P., Shi, Y.: Imagedream: Image-prompt multi-view diffusion for 3d
  generation. arXiv preprint arXiv:2312.02201  (2023)

\bibitem{wang2023pf-lrm}
Wang, P., Tan, H., Bi, S., Xu, Y., Luan, F., Sunkavalli, K., Wang, W., Xu, Z.,
  Zhang, K.: Pf-lrm: Pose-free large reconstruction model for joint pose and
  shape prediction. arXiv preprint arXiv:2311.12024  (2023)

\bibitem{wang2023slice3d}
Wang, Y., Lira, W., Wang, W., Mahdavi-Amiri, A., Zhang, H.: Slice3d:
  Multi-slice, occlusion-revealing, single view 3d reconstruction. arXiv
  preprint arXiv:2312.02221  (2023)

\bibitem{wang2004ssim}
Wang, Z., Bovik, A.C., Sheikh, H.R., Simoncelli, E.P.: Image quality
  assessment: from error visibility to structural similarity. TIP  (2004)

\bibitem{wu2023omniobject3d}
Wu, T., Zhang, J., Fu, X., Wang, Y., Ren, J., Pan, L., Wu, W., Yang, L., Wang,
  J., Qian, C., et~al.: Omniobject3d: Large-vocabulary 3d object dataset for
  realistic perception, reconstruction and generation. In: Proceedings of the
  IEEE/CVF Conference on Computer Vision and Pattern Recognition. pp. 803--814
  (2023)

\bibitem{xu2023dmv3d}
Xu, Y., Tan, H., Luan, F., Bi, S., Wang, P., Li, J., Shi, Z., Sunkavalli, K.,
  Wetzstein, G., Xu, Z., et~al.: Dmv3d: Denoising multi-view diffusion using 3d
  large reconstruction model. arXiv preprint arXiv:2311.09217  (2023)

\bibitem{yan2016perspective}
Yan, X., Yang, J., Yumer, E., Guo, Y., Lee, H.: Perspective transformer nets:
  Learning single-view 3d object reconstruction without 3d supervision.
  Advances in neural information processing systems  \textbf{29} (2016)

\bibitem{yang2022fvor}
Yang, Z., Ren, Z., Bautista, M.A., Zhang, Z., Shan, Q., Huang, Q.: Fvor: Robust
  joint shape and pose optimization for few-view object reconstruction. In:
  Proceedings of the IEEE/CVF Conference on Computer Vision and Pattern
  Recognition. pp. 2497--2507 (2022)

\bibitem{yao2018mvsnet}
Yao, Y., Luo, Z., Li, S., Fang, T., Quan, L.: Mvsnet: Depth inference for
  unstructured multi-view stereo. In: Proceedings of the European conference on
  computer vision (ECCV). pp. 767--783 (2018)

\bibitem{yu2022monosdf}
Yu, Z., Peng, S., Niemeyer, M., Sattler, T., Geiger, A.: Monosdf: Exploring
  monocular geometric cues for neural implicit surface reconstruction. Advances
  in neural information processing systems  \textbf{35},  25018--25032 (2022)

\bibitem{zhang2018lpips}
Zhang, R., Isola, P., Efros, A.A., Shechtman, E., Wang, O.: The unreasonable
  effectiveness of deep features as a perceptual metric. In: CVPR (2018)

\end{thebibliography}
\end{document}